# An Integrated Platform for Live 3D Human Reconstruction and Motion Capturing

Dimitrios S. Alexiadis, Anargyros Chatzitofis, Nikolaos Zioulis, Olga Zoidi, Georgios Louizis, Dimitrios Zarpalas, and Petros Daras, *Senior Member, IEEE*

*Abstract*—The latest developments in 3D capturing, processing, and rendering provide means to unlock novel 3D application pathways. The main elements of an integrated platform, which target tele-immersion and future 3D applications, are described in this paper, addressing the tasks of real-time capturing, robust 3D human shape/appearance reconstruction, and skeleton-based motion tracking. More specifically, initially, the details of a multiple RGB-depth (RGB-D) capturing system are given, along with a novel sensors' calibration method. A robust, fast reconstruction method from multiple RGB-D streams is then proposed, based on an enhanced variation of the volumetric Fourier transform-based method, parallelized on the Graphics Processing Unit, and accompanied with an appropriate texture-mapping algorithm. On top of that, given the lack of relevant objective evaluation methods, a novel framework is proposed for the quantitative evaluation of real-time 3D reconstruction systems. Finally, a generic, multiple depth stream-based method for accurate real-time human skeleton tracking is proposed. Detailed experimental results with multi-Kinect2 data sets verify the validity of our arguments and the effectiveness of the proposed system and methodologies.

## I. Introduction

3D RECONSTRUCTION of dynamic scenes, including human performers, and human motion tracking are important tasks in the fields of multimedia, computer vision, and graphics, with numerous applications, such as human motion analysis and recognition, dynamic 4D media exploration (e.g., in cultural heritage), mixed reality, and 3D telepresence/tele-immersion (TI). TI [1] refers to an emerging technology that can support realistic interpersonal communications, allowing geographically distributed users to share an activity in a common virtual space, where users are immersed via their real-time 3D replicant reconstructions.

Recent technological developments in the fields of real-time 3D capturing (e.g., Kinect and Tango), 3D displays and wearable 3D glasses (e.g., Oculus Rift and Microsoft Hololens), in combination with novel approaches for 4D (3D + time) content production, provide means to support novel applications, such as the above-mentioned ones. For example, recent advances in real-time capturing, full-3D reconstruction, and its compression [2] for transmission offer a technological basis to unlock novel 3D tele-immersive pathways.

This paper describes the main elements of an integrated platform, including capturing and fast 3D reconstruction of human 3D shape/appearance and skeleton-based motion tracking, which targets TI and future 3D applications. The elements of the continuously being developed platform have already allowed the realization of a number of relevant applications, as in http://vcl.iti.gr/3dTI/: 1) ski competition among users spread around Europe [3]; 2) 3D hang-out communications [4]; 3) multiplayer networked 3D games ("SpaceWars" and "Castle in the Forest"), where users participate via their on-the-fly reconstructed 3D replicants; and 4) athletes' training via professionals' performance capturing and reconstruction for quick-post 4D media. In addition, this paper describes a novel framework for the objective evaluation of the 3D reconstruction process, where the 3D ground-truth model is not available, as in real-time reconstruction applications.

This paper is organized as follows. In Section I-A, relevant existing work is given, prior to a summary of our contributions in Section I-B. Section II describes the employed multisensor 3D acquisition platform, its synchronization strategy, and a novel external calibration method. Sections III and IV provide the details of the proposed methods for 3D reconstruction and skeleton-based human motion tracking, respectively. The experimental results are presented in Section V, while the conclusions and future work are finally given in Section VI.

### A. Previous Relevant Work

*1) 3D Capturing and Reconstruction:* Several passive RGB camera-based reconstruction methods can be found in the literature [5]–[7]. With the exception of mainly shape-from-silhouette (SfS) (visual hull) methods [5], which lack the ability to reconstruct concavities and require a large number of cameras, unfortunately most methods are not applicable in our targeted real-time applications due to their slow performance. Other sophisticated human template-based reconstruction (performance capture) methods [8]–[10] are capable of generating temporally coherent 3D meshes using less cameras but still require a processing time of several minutes per frame.

Manuscript received October 2, 2015; revised February 20, 2016; accepted May 26, 2016. Date of publication June 7, 2016; date of current version April 3, 2017. This work was supported by the European Commission Project PATHway under Contract 643491. This paper was recommended by Associate Editor G. Thomas.

The authors are with the Information Technologies Institute, Centre for Research and Technology–Hellas, Thessaloniki 57001, Greece (e-mail: dalexiad@iti.gr; tofis@iti.gr; nzioulis@iti.gr; ozoidi@iti.gr; glouizis@iti.gr; zarpalas@iti.gr; daras@iti.gr).

This paper has supplementary downloadable material available at http://ieeexplore.ieee.org., provided by the author.

Color versions of one or more of the figures in this paper are available online at http://ieeexplore.ieee.org.

Digital Object Identifier 10.1109/TCSVT.2016.2576922

Regarding methods that use active direct-ranging sensors, explicit fusion [11] or volumetric implicit fusion methods [12] until recently had been applied only offline to combine range data from a single sensor. With the appearance of consumer-grade RGB-depth (RGB-D) cameras, variations of the referenced approaches have been employed for real-time telepresence applications [13]–[15]. In the category of volumetric reconstruction methods, the Poisson [16] and its ancestor Fourier transform (FT)-based reconstruction method [17], which require as input an oriented point set, are worth mentioning due to their robustness against noise in the input point-normal data. These methods, although fast, cannot perform in real time.

Real-time, full 3D (i.e., full body and 360°) reconstruction in this paper is achieved by capturing with multiple RGB-D sensors. An efficient multisensor telepresence/TI framework has been described in [18] and [19], that when compared with our approach (that reconstructs a single 360° full-3D mesh), combines multiple RGB-D data only at the rendering stage to produce intermediate views in a multiview depth image-based rendering framework. A single 3D mesh is reconstructed in [15] using a signed-distance-based volumetric method [12]. A similar reconstruction approach is described in [14]. Our proposed platform utilizes a volumetric approach, an enhanced variation of the FT-reconstruction method [17], which is parallelized on the Graphics Processing Unit (GPU) to achieve real-time reconstruction rates. Finally, the very recent and promising dynamic fusion work [20] has to be mentioned, which constitutes an extension of the known Kinect fusion system [21] and can reconstruct slowly deforming objects in a real-time SLAM framework. Nevertheless, it has been applied with a single handheld RGB-D sensor at close distances and does not address the texture-mapping problem.

Although the 3D processing methods proposed in this paper apply also to Kinect v1 data, the increased quality offered by the new Kinect generation v2 [22] naturally led to its adoption in our platform, which currently supports both versions. A few works regarding capturing with multiple Kinects v2 can be found in the literature [23], [24]. In a work [23] slightly relevant to ours, Kinects are used to capture a static room-sized scene into a virtual 3D model for safe testing of robot control programs. In that paper, however, neither the automatic calibration of the sensors nor the real-time capturing and reconstruction of dynamic scenes is addressed. In a very recent, more relevant work [24], a multi-Kinect2 capturing platform and its calibration are described. A server–client distributed capturing system is proposed, where the clients capture and on-the-fly generate and filter the raw 3D point clouds, which are either locally stored or transmitted to server, and are uncompressed in both cases. The software of a similar system can be found in the Brekel toolset (http://brekel.com/multikinectv2/). In contrast, the multiple Kinect RGB-D data in our system are compressed (before locally stored or on-the-fly transmitted), increasing the data rates and, more importantly, the data are fused to generate a single watertight and manifold textured mesh.

*2) 3D Reconstruction Evaluation:* Early studies on 3D reconstruction systems focused on reconstructing the surface of static objects under which the availability of the ground-truth 3D model is possible. Consequently, the reconstruction evaluation can be performed based on a 3D closest point framework or 3D Hausdorff distance when compared against the ground-truth model. However, recent 3D reconstruction systems perform fast 3D reconstruction of dynamic scenes captured from a single or from multiple RGB-D cameras [13], [15]. In such a case, the ground-truth model is not available, and the reconstruction evaluation is performed mainly subjectively. To the best of our knowledge, in this paper we propose the first framework for the objective evaluation of real-time 3D reconstruction systems.

*3) 3D Motion Capturing:* The most accurate solutions for human motion tracking are marker-based ones, which are intrusive and require special and expensive equipment, making them prohibitive in many practical applications. Marker-less solutions that use a depth sensor are mainly based on human motion databases and machine learning algorithms, enabling reliable human motion tracking by constraining the body configuration space [25], [26]. Kinect user-tracking is based on [26], but due to the one-side field-of-view, it is often problematic in challenging cases due to self-occlusions.

On the other hand, the proposed platform offers a generic method for accurate, real-time skeleton extraction based on a volumetric human representation (fully exploiting the information offered from the 3D reconstruction), thus overriding the problems in self-occlusion cases. Relevant approaches that try to extract the skeleton from volume data use mainly multiple RGB cameras and SfS algorithms [27]. Nevertheless, such approaches require robust silhouette extraction, which is not always an easy task, especially when the background is not static and uniform, thus introducing errors in the motion capturing method. Another similar method is proposed in [28], where the normalized gradient vector flow is extracted, based on partial differential equations. Likewise, in [29], Laplacian contraction is applied to skeletonize the volume. Although both methods offer a reliable skeleton, the identification and position estimation of the joints are not addressed because the latter is prohibitive for real-time applications. Finally, Straka *et al.* [30] utilize again an SfS algorithm and, similar to the proposed method, use graph-based techniques to detect and extract skeleton in real time. However, the experimental results provide only a coarse view of the method's accuracy. In particular, the experimental results present the success rate in joint position estimation, where as successful estimation is defined the case in which the extracted position is within a 100 mm radius from the ground truth. Contrary to that, the reliability and the accuracy of our method are demonstrated by comparing the anthropometric angle of the knee and elbow joints (flexion/extension) with the state-of-the-art marker-based motion capturing systems in challenging performances of the traditional sport skills.

### B. Summary of Contributions

This paper makes the following major contributions.
1) A multi-Kinect2 system, with distributed capturing and centralized processing nature. To the best of our

knowledge, this is among the first works in the literature regarding capturing with multiple Kinect 2 sensors.
2) A novel and fast sensors' calibration method.
3) A real-time reconstruction method from multiple RGB-D streams, which includes an enhanced variation of the volumetric FT-reconstruction method [17], parallelized on the GPU, and accompanied with appropriate texture mapping.
4) A novel framework for the quantitative evaluation of real-time 3D reconstruction systems from multiple RGB-D streams. To the best of our knowledge, this is the first work in the literature toward the specific objective.
5) A generic method for robust real-time skeleton extraction and tracking from multiple depth streams.
6) Finally, a large data set of multiple synchronized and calibrated RGB-D streams, captured with the proposed system and offered to the research community for experimentation.

## II. CAPTURING SETUP AND CALIBRATION

Although the proposed external calibration and 3D processing methods apply well with Kinect v1, in this section, we focus on the proposed multi-Kinect2 setup.

### A. Capturing System

The proposed system theoretically supports an arbitrary number of sensors, practically limited by the system's complexity, the needed processing power, and the local network bandwidth. In addition, one has to consider that interference between multiple Kinect v2 sensors exists, although it is less evident and its nature is completely different from that of Kinect v1. For a detailed analysis, see [31]. The devices are placed on a circle of radius $r \in [2\,m, 4\,m]$, all pointing to the center of the captured area (see also Fig. 2). As a good compromise between the system's complexity and coverage area, the use of $K = 4$ Kinect sensors is proposed. With careful placement of the devices, no major interference issues are observed.

Since Kinect2 limits its usage to one sensor per computer, a network architecture is mandatory. Specifically, the system uses $K$ computers, where $K - 1$ of them serve as slave nodes, and the remaining one serves as both slave and master node. The captured data arrive at the master node, where processing takes place. The platform operates in two different modes: a real-time mode and a quick-post one. The first mode continually polls the slave nodes/sensors for new synchronized (up to half of the sensors' internal clock interval) captured data. The latter mode signals all connected nodes to start recording data and finally triggers a gathering operation, having all nodes transfer recorded data to the master node. To enable fast transmission in the real-time mode and efficient storage in the quick-post one, an intra-frame compression scheme (JPEG for the RGB and LZ4 entropy compression for the depth images) was employed due to its reduced complexity and processing time. These modes enable either: 1) online 3D reconstruction, thus making it suitable for TI applications when combined with real-time efficient data compression [2] or 2) temporally complete and higher quality results, exploiting all the recorded data in a quick-post processing step.

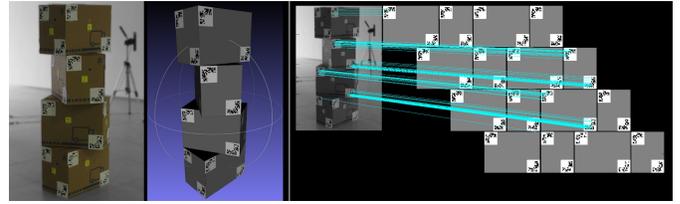

Fig. 1. From left to right: calibration structure, that was constructed using four standardized IKEA boxes (JÄTTENE, 600.471.51) along with 32 unique QR markers [ISO/IEC 18004:2006], its digitized 3D counterpart, and unwrapped texture image $M(\mathbf{u})$, and SIFT correspondences.

### B. Calibration

*1) Internal Calibration:* The Kinect2 depth-to-color mapping operation cannot be expressed as a fixed table since it depends on the depth measurements. Due to the centralized nature of processing the data, such a mapping table should be transferred to the master node in each frame, which is highly inefficient. Therefore, the mapping is approximated by a fixed KRT matrix (intrinsics and relative pose of the RGB camera). The approximation is performed based on 3D-to-2D correspondences in a dense 3D grid, obtained by employing the Kinect2 SDK functionality.

*2) External Calibration:* Spatial (external) calibration of the sensors is achieved through a novel registration method, utilizing an easy-to-build calibration structure that serves as a registration anchor. The registration is performed separately for each sensor, with respect to that anchor, using an exact digital replica of the calibration object. The approach is based on the scale-invariant feature transform (SIFT) [32] and Procrustes analysis [33]. Apart from constructing the calibration structure once, no user intervention is needed in contrast to commonly used methods that require capturing of a moving target.

*a) Calibration object:* The design of the calibration object was dictated by the following requirements: 1) to be universally easily reproducible; 2) to exhibit unique texture patterns to support SIFT feature extraction and matching; and 3) to be sufficiently large so that estimation/optimization is not affected by noise/inaccuracies in feature extraction and matching. To address these requirements, the calibration structure is realized with four standardized IKEA package boxes, of size $56 \times 33 \times 41\,cm^3$, as well as 32 unique quick response (QR) markers of area $13 \times 13\,cm^2$ placed at the corners of the boxes' side faces. An illustration of the calibration structure is given in Fig. 1 (left). As shown, the exact virtual counterpart of the calibration structure, a Computer-Aided Design (CAD) 3D model, was also designed. A complete manual with instructions and the CAD 3D model can be downloaded from http://vcl.iti.gr/3dTI/TCSVT. The virtual model's texture is unwrapped into a single image $M(\mathbf{u})$. Let $\mathcal{V}_M$ denote the set of the model's vertices and $\mathcal{S}_M$ the corresponding texture coordinates in $M(\mathbf{u})$.

*b) Calibration procedure:* The calibration structure is positioned at approximately the center of the capturing space so that: 1) the full object is in the field-of-view of all sensors and 2) in the case of Kinect v1, that eliminates any bias, the noise increases with the measured surface's distance.

A color image $I_k(\mathbf{u})$ and an accumulated depth image $\hat{D}_k(\mathbf{u})$ are acquired for each viewpoint $k$. During this step, to avoid any potential multi-Kinect interference issues, the sensors do not operate concurrently because the calibration object is static. To remove outliers and reduce noise, the accumulated depth image is obtained as the pixelwise median.

SIFT features are extracted from each color image $I_k(\mathbf{u})$, and feature correspondences are established with the *a priori* calculated features of the model texture image $M(\mathbf{u})$. Let $\mathbf{p}_k^j \leftrightarrow \mathbf{p}_M^j$, $j = 1, \ldots, J$ denote the $j$th established correspondence between the feature vector $\mathbf{p}_k^j$ from image $I_k(\mathbf{u})$ and the corresponding vector $\mathbf{p}_M^j$ from $M(\mathbf{u})$. Let $\mathbf{u}_k^j$ and $\mathbf{u}_M^j$ also denote the corresponding 2D image coordinates of the matched features. Given the $k$th sensor's intrinsic parameters and the depth image $\hat{D}_k(\mathbf{u})$, the 2D points $\mathbf{u}_k^j$ are backprojected to obtain the 3D points $\mathbf{V}_k^j$. The unwrapped texture coordinates $\mathbf{u}_M^j$ are transformed to the 3D vertices $\mathbf{V}_M^j$ by finding their nearest neighbors in $\mathcal{S}_M$, and thus their corresponding vertex position in $\mathcal{V}_M$. Given the 3D correspondences $\mathbf{V}_k^j \leftrightarrow \mathbf{V}_M^j$, the partial Procrustes problem [33] (no scaling and reflection) is solved to estimate the 6 Degrees of Freedom pose matrix of the $k$th sensor, i.e., by minimizing the sum of squared distances $\sum_j ||(\mathtt{R}_k \mathbf{V}_k^j + \mathbf{t}_k) - \mathbf{V}_M^j||^2$, subject to $\mathtt{R}_k^\mathrm{T} \mathtt{R}_k = \mathtt{I}_{3 \times 3}$.

### C. Quick-Post Synchronization

To synchronize the data recorded during the quick-post operation mode, a postsynchronization procedure is employed. Each sensor continuously acquires pairs of timestamped depth and color images. While it is not the exact case, the depth and color components are considered to be synchronized (in practice, they are synchronized up to 16 ms) and therefore the depth timestamps are used. Each Kinect generates timestamps according to its local timeline $\mathbb{T}_k$. An audio synchronization scheme is used to place the local timelines onto a global one. Audio signals of specific duration are simultaneously recorded from each sensor. Let the audio signal from the $k$th Kinect be denoted by $\mathbf{A}_k(t)$. Its delay with respect to the reference Kinect $k_0$ is calculated by $\hat{d}_k = \arg\max \ (R_{k,k_0}(d))$, where $R_{k,k_0}(d)$ is the cross correlation of audio signals $\mathbf{A}_k(t)$ and $\mathbf{A}_{k_0}(t)$. From these delays, the audio timestamp offsets $\check{T}_k$ are obtained, which are used to place the local timelines $\mathbb{T}_k$ onto the reference one $\mathbb{T}$.

Let the RGB-D timestamps, which are synchronized to the global timeline, be denoted by $T_k(n), n = 1, \ldots, N_k$, where $N_k$ is the total number of frames in the $k$th sequence. Under nominal Kinect operation conditions, these timestamps are approximately uniformly spaced with a time step of 33 ms. In practice, however, it is observed that frame generation rate can fluctuate. Therefore, a local synchronization scheme is employed, which continuously selects new groups of RGB-D frames as follows. Let an RGB-D frame with timestamp $T_k(n)$ be denoted by $\mathbb{F}_k(n)$. A synchronized group of frames (GoF) at a time instance $m$ is denoted by $\mathbb{G}(m) = \{\mathbb{F}_1(n_1(m)), \mathbb{F}_2(n_2(m)), \ldots, \mathbb{F}_K(n_K(m))\}$. The synchronization inconsistency of a GoF is measured by the maximum timestamp difference of its frames, i.e., $\max_{i,j}\{|T_i(n_i(m)) - T_j(n_j(m))|\}$. Given a GoF $\mathbb{G}(m)$, in order to generate the next group $\mathbb{G}(m+1)$, all candidate combinations $\acute{\mathbb{G}}(m;s) = \{\mathbb{F}_1(n_1(m) + s_1), \mathbb{F}_2(n_2(m) + s_2), \ldots, \mathbb{F}_K(n_K(m) + s_K)\}$ are considered, where $\mathbf{s} = [s_1, s_2, \ldots, s_K] \in \{0,1\}^K$ is a binary string of length $K$ (excluding zero). Put simply, a new GoF is generated by moving in some or in all the timelines by one step. Among all candidates $\acute{\mathbb{G}}(m;s)$, the one that minimizes the synchronization inconsistency is selected. The algorithm continues iteratively until the end of a sequence.

## III. RECONSTRUCTION OF GEOMETRY AND APPEARANCE

The performance of a captured user along time is reconstructed by the extraction of the user's 3D geometry and appearance on a per-frame basis, i.e., for each time instance. Therefore, given multiple captured depth maps $D_k(\mathbf{u})$, $\mathbf{u} = (u, v)^\mathrm{T}$, $k = 1, \ldots, K$ at a specific time instance along with the corresponding RGB images, the objective is the fast 3D reconstruction in the form of a single textured triangular mesh.

Let $\mathbf{u} \leftarrow \Pi_k(\mathbf{X})$ define the world-to-projective mapping operation, which maps a 3D point $\mathbf{X} = (X, Y, Z)^\mathrm{T}$ to a pixel $\mathbf{u}$, while $\mathbf{X} \leftarrow \Pi_k^{-1}(\mathbf{u}, Z)$ denotes the inverse (projective-to-world) mapping. Similarly, let $\Pi_k^\mathrm{RGB}(\mathbf{X})$ stand for the corresponding mapping for the $k$th RGB camera.

### A. Raw Reconstruction and Confidence Weights

For each foreground pixel $\mathbf{u} \in \mathcal{F}_k$ on the $k$th depth map, a raw 3D point $\mathbf{X}_k(\mathbf{u}) = \Pi_k^{-1}(\mathbf{u}, D_k(\mathbf{u}))$ is reconstructed. We use the notation $\mathbf{X}(\mathbf{u})$ to highlight that each reconstructed 3D point $\mathbf{X}$ is associated with a foreground pixel $\mathbf{u} \in \mathcal{F}_k$ on the image plane. In addition, the corresponding raw 3D normals $\mathbf{N}_k(\mathbf{u})$ are estimated as follows. Terrain step discontinuity constraint triangulation [13] is used to realize an organized triangulation scheme in which each vertex may be connected to one of its eight neighbors (on the 2D image plane). Given the triangle normals, each vertex is assigned the mean of the normals of the triangles into which it participates.

Apart from the raw position-normal information, a confidence-weight map $W_k(\mathbf{u})$ is calculated on a per-vertex basis, based on the following intuitive observations. The quality of a raw measurement depends on the depth-camera's viewing angle, i.e., the angle between the camera's line-of-sight and the surface normal. Therefore, a confidence value for a pixel (vertex) $\mathbf{u} \in \mathcal{F}_k$ is computed from $W_{k,1}(\mathbf{u}) = \max\{\langle \hat{\mathbf{X}}_k^{\mathrm{loc}}(\mathbf{u}), \mathbf{N}_k^{\mathrm{loc}}(\mathbf{u}) \rangle, 0\}$, where $\hat{\mathbf{X}} = -\mathbf{X}/||\mathbf{X}||$, $\langle \cdot, \cdot \rangle$ denotes the inner vector product, and the superscript loc denotes that the 3D positions and normals are defined with respect to the local camera's coordinate system. In practice, the depth measurements near the foreground object's silhouette boundaries are noisy. An associated confidence map $W_{k,2}(\mathbf{u}) \in [0, 1]$ is extracted based on this observation. A fast approach to calculate such a confidence value for a specific pixel $\mathbf{u}$ is to count the number of foreground pixels inside a square neighborhood around $\mathbf{u}$, divided by the neighborhood's size. This is implemented efficiently using a 2D moving average filter (with radius 10 pixels in our experiments) on the corresponding binary silhouette image. The final confidence map is calculated from the product $W_k(\mathbf{u}) = W_{k,1}(\mathbf{u}) \cdot W_{k,2}(\mathbf{u})$.

### B. 3D Volume Reconstruction

The objective is to calculate a scalar volume function $A(\mathbf{q})$, which implicitly contains the surface information as the isosurface at an appropriate level $L$. The function is defined over a 3D grid $\mathbf{q} = [q_X, q_Y, q_Z]^T \in \{0, \ldots, N_X - 1\} \times \{0, \ldots, N_Y - 1\} \times \{0, \ldots, N_Z - 1\}$, inside the foreground object's bounding box. To this end, an FT-based approach [17] is employed and enriched with a smoothing and weighting scheme.

The raw normals $\mathbf{N}_k(\mathbf{u})$ are initially splatted to the voxel grid to obtain the gradient vector field $\mathbf{V}(\mathbf{q})$. In the simplest nonweighted version of the method, each raw sample is clapped to its nearest voxel and then the vector field is normalized by the number of samples clapped at each voxel. In the proposed method's version, the normal $\mathbf{N}_k(\mathbf{u})$ is smoothly distributed to point's neighbor voxels, according to

$$\mathbf{V}(\mathbf{q}) = \sum_k \sum_{\mathbf{u} \in \mathcal{F}_k} g(\mathbf{X}_k(\mathbf{u}), \mathbf{q}; \sigma_1) \cdot [w_k(\mathbf{u}, \mathbf{q}) \cdot \mathbf{N}_k(\mathbf{u})] \quad (1)$$

where $g(\mathbf{X}_k(\mathbf{u}), \mathbf{q}; \sigma_1)$ are the splatting weights based on the distance $x$ of point $\mathbf{X}$ from voxel $\mathbf{q}$, and more specifically, $g(x; \sigma_1) = \sigma_1^{-1} \exp(-x^2/\sigma_1^2)$ is a Gaussian. The confidence-related weights $w_k(\mathbf{u}, \mathbf{q})$ are obtained from: $w_k(\mathbf{u}, \mathbf{q}) = W_k(\mathbf{u})/d(\mathbf{q})$, with the normalization factor $d(\mathbf{q})$ being a weighted estimate of the points density at the voxel $\mathbf{q}$, namely

$$d(\mathbf{q}) = \sum_k \sum_{\mathbf{u} \in \mathcal{F}_k} g(\mathbf{X}_k(\mathbf{u}), \mathbf{q}; \sigma_2) \cdot W_k(\mathbf{u}) \quad (2)$$

where $g(x; \sigma_2)$ is again a Gaussian with standard deviation $\sigma_2$. In other words, we employ kernel density estimation [34] [considering the weights $W_k(\mathbf{u})$] using a Gaussian kernel. To avoid singularities, $\sigma_2$ should always be larger than $\sigma_1$. It was experimentally selected equal to $\sigma_2^2 = 3/2\sigma_1^2$. With respect to $\sigma_1$, the larger its value, the smoother the output gradient field and the reconstruction is expected to be. A reasonable selection is to use a $\sigma_1$ value that is proportional to the voxel's diagonal. In our experiments, we use a relatively small value, equal to voxel's radius (half of the diagonal). Given this selection, to speed up calculation in our implementation, we consider only the $4^3$ voxels around each input point since the values at other voxels will be very low.

Intuitively, the use of the splatting weights in (1) is similar (not equivalent) to convolving with a low-pass filter, resulting into a smooth gradient field. The use of the density-normalized weights $w_k(\mathbf{u}, \mathbf{q})$ assigns smaller weights to non-confident input samples at high-density regions, letting other confident points in the neighborhood contribute more in the reconstruction of the gradient field.

Subsequently, following [17], the calculated gradient field $\mathbf{V}(\mathbf{q}) = [V_X(\mathbf{q}), V_Y(\mathbf{q}), V_Z(\mathbf{q})]^T$ is transformed into the 3D frequency domain by applying 3D Fast Fourier Transform (FFT) separately to each of vector field's $X$, $Y$, and $Z$ components to obtain $\hat{\mathbf{V}}(\boldsymbol{\omega}) = [\hat{V}_X(\boldsymbol{\omega}), \hat{V}_Y(\boldsymbol{\omega}), \hat{V}_Z(\boldsymbol{\omega})]^T$, where $\boldsymbol{\omega} = (\omega_x, \omega_y, \omega_z)^T$ is the 3D frequency vector. The integration filter $\hat{\mathbf{F}}(\boldsymbol{\omega}) = [\hat{F}_X(\boldsymbol{\omega}), \hat{F}_Y(\boldsymbol{\omega}), \hat{F}_Z(\boldsymbol{\omega})]^T = (1/||\boldsymbol{\omega}||^2)[j\omega_x, j\omega_y, j\omega_z]^T$, $j = \sqrt{-1}$ is applied by multiplication in the frequency domain.

The final volumetric function $A(\mathbf{q})$ is calculated by applying the inverse 3D FFT on the integrated (filtered) vector field and adding its $X$, $Y$, and $Z$ components. The purpose of applying the integration filter in the frequency domain is justified by the infinite impulse response nature of the filter, which does not allow for parallel calculations on the GPU if applied in the original domain, as well as by the existence of very fast FFT implementations.

It has to be recalled here that multiplication in the discrete Fourier domain is equivalent to circular convolution in the original spatial domain. Therefore, to avoid any unwanted effects of circular convolution, the tight foreground object's bounding box is adequately extended before voxelization [equivalent to zero padding of $\mathbf{V}(\mathbf{q})$].

The final 3D surface is extracted in the form of a triangle mesh (vertex positions, normals, and connectivity), as the isosurface $A(\mathbf{q}) = L$ using the marching cubes algorithm [35]. The level $L$ is calculated as the average value of $A(\mathbf{q})$ at the input sample locations $\mathbf{X}_k(\mathbf{u})$. The whole reconstruction method was implemented with Compute Unified Device Architecture (CUDA) (www.nvidia.com/object/cuda_home_new.html) for parallel computing on the GPU, since most of its stages involve pixelwise or voxelwise calculations.

### C. Texture Mapping—Reconstruction of Appearance

Many vertices in the final reconstructed model are visible in more than one RGB cameras. Therefore, colors from more than one RGB camera have to be combined to produce the color of each reconstructed vertex. There are two important issues that need to be considered and can significantly improve the visual quality of the rendered reconstruction. First, volumetric 3D reconstruction methods generally produce a relatively low number of triangles and vertices (depending on the volume resolution), lower than the number of pixels in the original 2D domain. Therefore, a color-per-vertex rendering approach will lead to color aliasing, producing low visual quality. Instead, we employ full texture mapping and assign multiple texture patches to each triangle from the multiple RGB views. Second, instead of using equal weights for each visible RGB camera, one could use weights based on the quality of the captured colors. Practically, given that the RGB cameras are more-or-less equidistant from the captured user.

1) The quality of the captured color depends on the viewing angle of the captured surface, i.e., it depends on the angle between the line-of-sight and the surface normal.
2) Near the captured object boundaries, inaccurate depth-to-RGB camera registration (calibration) may lead to color-mapping artifacts (e.g., color of the background assigned on the reconstructed foreground object).

Therefore, the captured color information near the object boundaries has to be assigned a smaller weight. It should be noted that the depth image-based weights $W_k(\mathbf{u})$ in Section III-A were defined based on similar intuitions. Finally, given that the depth and RGB cameras of a single Kinect-like RGB-D device are parallel and very close to each other, it can be practically considered that the visibility of a vertex is the same in both cameras. In practice, this approximation proved to be helpful in speeding up calculations without introducing significant color artifacts. Since the weights $W_k(\mathbf{u})$ contain visibility information and incorporate the practical observations for weighting, they are directly used in the texture-mapping process.

Formally, let $\mathcal{V}(\mathbf{X}) \subseteq \{1, \ldots, K\}$ denote the subset of depth cameras in which the vertex $\mathbf{X}$ is visible. Let $\mathbf{u}_k, k \in \mathcal{V}(\mathbf{X})$ also be the corresponding pixels on the visible depth cameras, where the vertex $\mathbf{X}$ projects according to $\mathbf{u}_k = \Pi_k(\mathbf{X})$. Similarly, let $\mathbf{u}_k^{\text{RGB}} = \Pi_k^{\text{RGB}}(\mathbf{X})$ be the corresponding pixels (UV coordinates) on the visible RGB cameras. Each vertex is assigned multiple weights $W_k(\mathbf{u})$ and UV-texture coordinates $\mathbf{u}_k^{\text{RGB}}$ on the corresponding visible images. Each reconstructed triangle is rendered with OpenGL multitexture blending using the associated vertices' weights.

### D. Color Correction

The RGB cameras of consumer-grade sensors, especially under nonuniform lighting and background conditions, may output color values that vary significantly between adjacent RGB views, i.e., the color of the same 3D point appears different in two captured RGB views. To attenuate the resulting texture artifacts, we search for the color-correction functions that minimize (in a robust mean-square sense) the color difference between the pairs of pixels in two cameras that capture (approximately) the same 3D point. Our approach borrows ideas from [18] but uses the Hue-Saturation-Value (HSV) color space instead.

*1) Searching for Color Correspondences:* Consider two adjacent RGB cameras with overlapping field-of-view and indexed with $k_1$ and $k_2$. Let the raw vertex positions for a given frame be denoted by $\mathbf{X}_{k_1}^i, i = 1, \ldots, I$ and $\mathbf{X}_{k_2}^j, j = 1, \ldots, J$, respectively, whereas the corresponding raw RGB vertex colors be $\mathbf{C}_{k_1}^i$ and $\mathbf{C}_{k_2}^j$. The mutual closest points between the point clouds with Euclidean distance smaller than 20 mm are searched. This way, a number of color correspondences $\mathbf{C}_{k_1}^m \leftrightarrow \mathbf{C}_{k_2}^{n(m)}, m = 1, \ldots, M$, are found. To achieve robustness, color correspondences in multiple frames are accumulated.

*2) Estimating Color-Correction Functions:* The objective is to find a linear function $\mathbf{F}_{k_1,k_2}(\mathbf{C})$, such as $||\mathbf{F}_{k_1,k_2}(\mathbf{C}_{k_1}^m) - \mathbf{C}_{k_2}^{n(m)}||$, is minimized. We found in practice that an RGB-separately approach [18] may be ill-posed when the range of colors in the foreground object is limited, e.g., when a specific color channel is missing. On the other hand, by working in the HSV color space, it is expected that the Hue component is not affected by the exposure control, while it was experimentally found that the saturation component is only slightly affected. Therefore, the correspondence colors are transformed into the HSV color space, and a linear mapping model is built by robust (RANSAC) linear regression on the value data $V_{k_1}^m$ and $V_{k_2}^{n(m)}, m = 1, \ldots, M$, such that $|F_{k_1,k_2}(V_{k_1}^m) - V_{k_2}^{n(m)}|$ is minimized in a (robust) mean-square sense. Given a reference camera, the final color-correction function for a specific camera is obtained by considering the path from that camera to the reference.

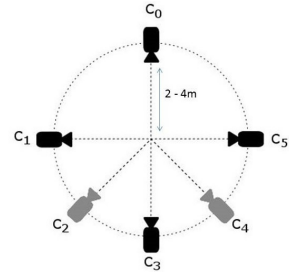

Fig. 2. Camera setup for 3D reconstruction and quantitative evaluation. $K = 4$ sensors ($c_0$, $c_1$, $c_3$, and $c_5$) take part in the reconstruction process, and $K' = 2$ sensors ($c_2$ and $c_4$) serve as additional ground-truth views.

### E. Methodology for Quantitative Objective Evaluation

To objectively evaluate the performance of a real-time 3D reconstruction method, a capturing system consisting of $K + K'$ calibrated RGB-D sensors is employed whereby $K$ sensors take part in the reconstruction procedure, and $K'$ sensors serve as additional ground-truth planar views of the user. Such a capturing system is shown in Fig. 2. The proposed objective evaluation framework aims at addressing the following question. How well does the reconstructed mesh explain (match) the captured data in all available 2D views? The comparison is performed on the 2D image plane by: 1) projecting the reconstructed mesh into the $K + K'$ planar views and 2) comparing the rendered depth and color images (the depth and color buffers of the OpenGL frame buffer) with the original captured views. Since the ground-truth data are sensor data, they may suffer from noise, especially near the border areas between the user and the background. This means that the employed performance measures may sometimes deteriorate due to bad ground-truth model assumption. Such situations will be further discussed in Section V.

*1) Evaluation of the Reconstructed Volume:* The reconstructed 3D shape may suffer from holes, missing/cut limbs, and model distortions. To quantify such errors, the percentage of the nonreconstructed object volume is estimated based on the silhouette information as follows. First, the reconstructed 3D mesh is projected onto the depth image plane of sensor $c_k$, $k = 1, \ldots, K + K'$, and the reconstruction's binary 0/1 silhouette mask $S_k^r$ is extracted. The ground-truth silhouette mask $S_k^g$ is also extracted by performing foreground–background segmentation on the sensor depth. The volume reconstruction error (VRE) metric is calculated from

$$V_k = |S_k^r \oplus S_k^g| / |S_k^r \vee S_k^g| \quad (3)$$

where $\oplus$ and $\vee$ denote the binary operators Xor and Or, respectively, and $|\cdot|$ is the silhouette mask area. Due to the Xor operation, the metric punishes both false positive and false negative silhouette areas. Another used silhouette-based metric is the 2D Hausdorff distance [36], expressed in pixels

$$H_k = \max\left\{\sup_{\mathbf{u}_r \in S_k^r} \inf_{\mathbf{u}_g \in S_k^g} d(\mathbf{u}_r, \mathbf{u}_g), \sup_{\mathbf{u}_g \in S_k^g} \inf_{\mathbf{u}_r \in S_k^r} d(\mathbf{u}_r, \mathbf{u}_g)\right\} \quad (4)$$

where $d(\mathbf{u}_r, \mathbf{u}_g)$ denotes the 2D distance between the pixel $\mathbf{u}_r$ in the reconstructed silhouette mask and the pixel $\mathbf{u}_g$ in the ground-truth mask. When the reconstructed model contains holes, the Hausdorff distance is equal to the radius of the circle inscribed to the hole. When it contains a missing or cut limb, the metric will be equal to the length of that limb.

*2) Evaluation of the Reconstructed Geometry:* To evaluate how accurately the 3D geometry is reconstructed, a 3D closest point approach is employed. First, the ground-truth foreground depth image of the sensor $c_k$ is backprojected onto the 3D space to generate a point cloud $\{\mathbf{X}_{k,i}^g, i = 1, \ldots, I_k\}$. The point cloud $\{\mathbf{X}_{k,j}^r, j = 1, \ldots, J_k\}$ is also generated from the corresponding depth image obtained from the reconstructed mesh. The use of a closest point rooted mean-square error (CP-RMSE) metric is proposed, given from

$$\text{CPRMSE}_k = \sqrt{\frac{1}{I_k} \sum_{i=1}^{I_k} \inf_{j=1,\ldots,J_k} \{\|\mathbf{X}_{k,i}^g - \mathbf{X}_{k,j}^r\|^2\}}. \quad (5)$$

A closest point MSE metric is employed instead of 3D Hausdorff distance between surfaces, since the latter would require connectivity information for the ground-truth point cloud and, more importantly, due to its sup operation (instead of mean), it would mainly count for missing limbs (as in Section III-E1), instead of the reconstruction geometry accuracy.

*3) Evaluation of the Appearance Quality:* The evaluation of the appearance quality is perceived as an image quality assessment task. How well does the ground-truth RGB image, captured from a specific viewpoint, match the textured model, rendered from exactly the same viewpoint? Due to the poor performance of MSE and Peak Signal-to-Noise Ratio as visual quality metrics [37], a structural similarity (SSIM) index-based measure [38] was chosen in our framework. The SSIM between two images, evaluated at pixel $\mathbf{u}$, is given from $\text{SSIM}(\mathbf{u}) = [l(\mathbf{u})]^\alpha \cdot [c(\mathbf{u})]^\beta \cdot [s(\mathbf{u})]^\gamma$, where $\alpha$, $\beta$, and $\gamma$ are the constant exponents, and

$$l(\mathbf{u}) = \frac{2\mu_X \mu_Y + C_1}{\mu_X^2 + \mu_Y^2 + C_1}, \quad c(\mathbf{u}) = \frac{2\sigma_X \sigma_Y + C_2}{\sigma_X^2 + \sigma_Y^2 + C_2}$$
$$s(\mathbf{u}) = \frac{2\sigma_{XY} + C_3}{\sigma_X \sigma_Y + C_3} \quad (6)$$

are the luminance, contrast, and structural terms, respectively. $C_1$, $C_2$, and $C_3$ are small constants and $\mu_X$, $\mu_Y$, $\sigma_X$, $\sigma_Y$, and $\sigma_{XY}$ stand for the images' means, standard deviations, and cross covariances in a neighborhood $\mathcal{N}(\mathbf{u})$ around pixel $\mathbf{u}$. In this paper, a variation of SSIM is employed, the weighted multiscale SSIM (WMS3IM) [39] is evaluated at $J = 3$ scales. WMS3IM is calculated from $\text{WMS3IM}_k(\mathbf{u}) = \prod_{j=1}^{3} l_j^{\alpha_j} c_j^{\beta_j} s_j^{\gamma_j}$, where $j = 1, 2, 3$ stands for the scale and the constants $\alpha_j, \beta_j, \gamma_j$ have been set based on the psychovisual experiments of [40], and more specifically $\{\alpha_1, \alpha_2, \alpha_3\} = \{0, 0, 0.1333\}$ and $\{\beta_1, \beta_2, \beta_3\} = \{\gamma_1, \gamma_2, \gamma_3\} = \{0.0448, 0.3001, 0.1333\}$. The structural term at scale $j$ is calculated from

$$s_j = \frac{\sum_{\mathbf{u} \in S_k^r} s(\mathbf{u}) w(\mathbf{u})}{\sum_{\mathbf{u} \in S_k^r} w(\mathbf{u})} \quad (7)$$

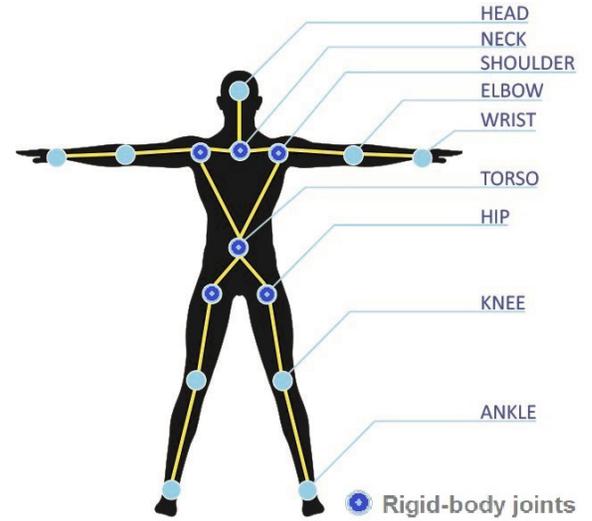

Fig. 3. 15-joint structure, separated into the rigid-body part and the limbs part.

where the weights in our case are $w(\mathbf{u}) = \sum_{\mathbf{v} \in \mathcal{N}(\mathbf{u})} S_k^r(\mathbf{v})$. Similar equations are used to calculate the luminance and contrast terms $l_j$ and $c_j$.

## IV. VOLUME-BASED MOTION TRACKING

In this section, a fast method for human skeleton tracking is presented, exploiting the human volume reconstructed as in Section III. The method tracks the joint positions of a 15-joints skeletal structure, as shown in Fig. 3. This structure is separated into: 1) the rigid-body part that includes the torso, hip, neck, and shoulder joints and 2) the limb parts that consist of the elbow and wrist or the knee and ankle joints. The rigid-body part is a group that moves rigidly based on the assumption that the relative rotations of the upper- and lower-body trunk can be ignored. This simplification, although it constitutes a limitation, introduces robustness.

The proposed method consists of two phases. Initially, in a user-calibration phase, the user body structure is estimated. Then, during the main tracking phase, both the position orientation of the rigid-body part and the limb-joint positions are tracked. The main tracking algorithm is initially described, assuming that the necessary user-calibration data are known, before going into the description of the user-calibration phase in Section IV-B. The algorithm steps that are performed on a per-frame basis are given sequentially in Section IV-A.

### A. Main Tracking Algorithm

*1) Volume Binarization and Skeletonization:* Given the reconstructed volume function $A(\mathbf{q})$ and the corresponding isosurface level $L$ (Section III-B), the binary human volume $A_h(\mathbf{q}) \in \{0, 1\}$ is extracted [Fig. 4(a)]. Skeletonization is then realized [Fig. 4(b)], using the method in [41]. The result is denoted by $A_s(\mathbf{q})$. In addition, we let $Q_h$ denote the set of voxels belonging to the binary volume $A_h$, i.e., $Q_h = \{\mathbf{q} :$

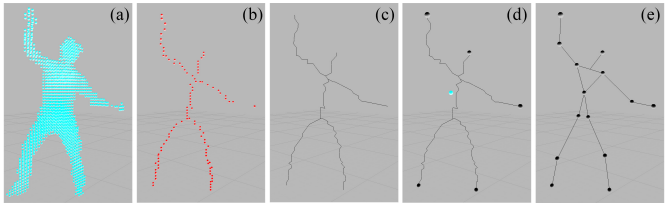

Fig. 4. Overview of the main stages of the proposed method. (a) Initial binary volume $A_h$. (b) Skeletonized volume $A_s$. (c) MST of graph $\mathcal{G}(\mathcal{V}_s, \mathcal{E})$. (d) Six basic/initial joints detection. (e) Final extracted skeleton.

$A_h(\mathbf{q}) = 1\}$, and $Q_s$ the voxels belonging to the skeletonized volume, respectively.

*2) Estimation of Torso Position:* Given the structure and symmetry of the human body, the torso is in most cases the joint closest to the human mass center [42]. Therefore, the most centralized voxel of $Q_h$ is initially searched. More specifically, let $\mathbf{p}(\mathbf{q})$ denote the 3D coordinates of voxel $\mathbf{q}$. The average Euclidean distance of a voxel $\mathbf{q}$ with the rest voxels is $D(\mathbf{q}) = (1/|Q_h|) \sum_{\mathbf{q}_i \in Q_h} \|\mathbf{p}(\mathbf{q}_i) - \mathbf{p}(\mathbf{q})\|$, where $|Q_h|$ is the cardinality of $Q_h$. The point $\mathbf{p}(\mathbf{q}_c)$ for which $D(\mathbf{q})$ is minimized represents the voxel closest to the torso. The point in the set $Q_s$, closest to $\mathbf{p}(\mathbf{q}_c)$, represents the detected torso position and is notated as $\mathbf{p}_t$.

*3) Detection of Extreme Joints (Head, Wrists, and Ankles):* Toward our objective, a graph-based technique is utilized.

1) *Graph and Minimum Spanning Tree (MST):* The points of the skeletonized set $Q_s$ are considered as the vertices (nodes) $\mathcal{V}_s$ of a graph $\mathcal{G}(\mathcal{V}_s, \mathcal{E})$, where $\mathcal{E}$ is the edge set. The graph is constructed by connecting the nodes with Euclidean distance lower than a predefined radius (i.e., ~15 cm) so that only neighboring vertices are connected. The cost of an edge between two connected nodes is set equal to their Euclidean distance. The cost along a path from one node to another equals their geodesic distance. The MST, let $\mathcal{T}$, is extracted from $\mathcal{G}$, using Kruskal's algorithm [43], as shown in Fig. 4(c). The MST provides an initial skeleton-like model, with unique paths from node to node.

2) *Extreme Joints Detection:* Exploiting the structure of the given MST under normal circumstances its leaves correspond to the human-body extremities, as shown in Fig. 4(d). These five extremities need to be labeled as ankle, wrist, or head.

However, in special cases, the leaves of the initial tree $\mathcal{T}$ may not count to $N = 5$. Let $N_d$ denote the number of the MST's leaves. In the nonstandard case of $N_d < 5$, indicating possible body part stacking, a heuristic approach is used. The two lower detected leaves (their 3D positions have the lowest values along the y-axis) are labeled as ankle joints. Given that, let $\mathcal{T}_{\text{low}}$ denote the subtree that includes the paths from the ankles to the torso. Subtracting $\mathcal{T}_{\text{low}}$ from $\mathcal{T}$ (i.e., dropping the nodes of $\mathcal{T}_{\text{low}}$ and their incident edges), the upper-body subtree $\mathcal{T}_{\text{up}} = \mathcal{T}/\mathcal{T}_{\text{low}}$ is obtained. In the cases of holding the hands stacked on the body, the number of the leaves of $\mathcal{T}_{\text{up}}$ will be equal to 3, i.e., the wrists joints are revealed. In the other nonstandard rare case of $N_d > 5$, indicating spurious artifact limbs, the detected leaves are filtered based on their geodesic distance to the torso. The paths with geodesic lengths closer to those estimated during the calibration phase are selected, while the rest of them are dropped. Thus, the number of the leaves in the final tree equals to $N = 5$.

Let $\mathcal{H} = \{H_p\}_{p=1,\dots,10}$ denote the set of all paths from extremity to extremity. Let $\mathcal{B} = \{B_p\}_{p=1,\dots,6}$ also denote the subset of $\mathcal{H}$ that includes only the paths passing through the torso point $\mathbf{p}_t$, i.e., the paths from an upper-body extremity to a lower-body one. The intersection of the paths in $\mathcal{B}$ (i.e., keeping only the nodes common in these paths) gives the spine path $S$. The detection of the spine path is crucial, since its usage is twofold: 1) it separates the extremities into the upper-body (wrists and head) and lower-body (ankles) groups. The upper-body joint with the shortest path to the torso is labeled as the head and 2) the torso orientation can be estimated by applying Principal Components Analysis (PCA) on the area around the spine.

*4) Torso Orientation Estimation and Rigid-Body Update:* Given the spine path $S$ and the initial volume $A_h$, we extract the points in the area of the thorax, the abdominal, and the pelvis segments, by considering a radius $r$ whose value depends on the human-body volume, around the points of the spine. Let the set of these points be $\mathcal{P}_{\text{tr}}$. By applying PCA to $\mathcal{P}_{\text{tr}}$, the torso orientation $\mathrm{R}_t$ is estimated. Assuming that the neck, the shoulders, the hips, and the torso are rigidly connected (rigid-body part), we use $\mathrm{R}_t$ and the torso position $\mathbf{p}_t$ to transform the root-rigid body in the world space.

*5) Detection of Link Joints (Elbows and Knees):* Let $\mathbf{X}_r$ and $\mathbf{X}_x$ stand for the position of the root joint (i.e., hip or shoulder) and the corresponding human extremity (i.e., wrist or ankle) of a limb, respectively. Let $\mathbf{X}_j$ also be the position of a node along the path from $\mathbf{X}_r$ to $\mathbf{X}_x$. The bone lengths are considered to be known, estimated during the user-calibration phase. The positions of the link joints are extracted from

$$\hat{\mathbf{p}} = \arg\min_j (|\|\mathbf{X}_j - \mathbf{X}_r\| - d_r| + |\|\mathbf{X}_j - \mathbf{X}_x\| - d_x|) \quad (8)$$

where $d_r$ is the bone length from joint $r$ to $j$, and $d_x$ is the length from joint $x$ to $j$. This means that $\hat{\mathbf{p}}$ is given as the point on the skeleton graph that intersects with the circular patch obtained from the intersection of the spheres $\{\mathbf{X}_r, d_r\}$ and $\{\mathbf{X}_x, d_x\}$.

*6) Kalman Filtering:* Kalman filtering [44] is applied on a per-joint basis to achieve a smooth transition from frame to frame and avoid errors from volume noise. Erroneous estimates of joint positions (especially under circumstances like self-occlusion or ghost limbs) can be partially corrected by imposing inter-frame correlation of joint positions via Kalman filtering. In the employed Kalman filter model, the state transition matrix is set based on the Newtonian law $\mathbf{p}(t) = \mathbf{p}(t-1) + \mathbf{v}(t-1)$, while the measurement/observation vector corresponds to the estimated 3D joint position and is modeled as the actual position plus zero-mean Gaussian white noise.

### B. Human-Body Structure Calibration

The calibration phase assumes that the user is standing in X-pose, as shown in Fig. 5(a). Initially, the rigid-body part structure is estimated, as shown in Fig. 5(b). The bounding box of the body trunk is extracted, and the intersections of the edges of the MST with this bounding box give the shoulder and hip joints. The neck is extracted as the midpoint between the shoulders. The positions of the link joints (elbows and knees) in the $X$-pose [Fig. 5(a)] are then extracted from

$$\hat{\mathbf{p}} = \arg\max_j \left( \frac{||(\mathbf{X}_j - \mathbf{X}_r) \times (\mathbf{X}_x - \mathbf{X}_r)||}{||(\mathbf{X}_x - \mathbf{X}_r)||} \right). \quad (9)$$

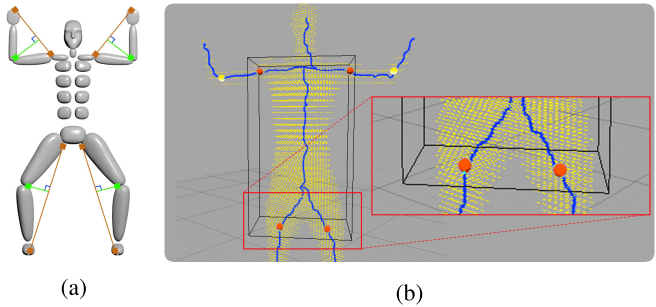

(a) (b)

Fig. 5. User calibration in $X$-pose. (a) Estimation of bone lengths. (b) Extraction of the rigid-body structure.

The notation is similar to the one in (8). According to (9), based on human-body bone rigidness, the point with the maximum distance from the line segment that connects the joints (e.g., shoulder with wrist) represents the link-joint position, as shown in Fig. 5(a). During the calibration phase, apart from the bone lengths, the geodesic lengths of the paths from each joint to torso are extracted.

The method is applied for a sequence of frames instead of a single frame. The body structure definition is considered complete after a few frames in which the rules of human-body symmetry and estimation repeatability were satisfied.

## V. Experimental Results

We initially present the results of the employed capturing and reconstruction method, in terms of subjective 3D geometry/appearance reconstruction quality and processing time, before going into an objective quantitative evaluation analysis, which is based on the proposed framework of Section III-E. In Section V-C, the experimental results of the proposed human skeleton-tracking method are finally presented.

Additional experimental results, in the spirit of this section, can be found in the Supplementary Material, along with Supplementary Videos. The data sets used in this section can be downloaded from http://vcl.iti.gr/3dTI/TCSVT/dataset.

### A. 3D Reconstruction Results and Processing Time

Most results were obtained using capturing setups with multiple Kinects2, in both small-area and medium-area spatial configurations. In the second case, professional athletes are captured performing skills of traditional Gaelic and Basque sports. The presented results were obtained using a volume resolution $2^r \times 2^{r+1} \times 2^r$ with $r = 7$, unless otherwise stated. Notice that the resolution along $Y$ is doubled, as the human bounding box is larger along its height.

*1) Small-Area Configuration:* Four sensors are placed on a circle of radius approximately 2.5 m with an individual performing athletic movements at the center of the captured space.

*a) Argyris sequence:* In Fig. 6(a), the proposed reconstruction result is compared with the initial reconstructed data (four aligned separate meshes) in terms of 3D geometry. Despite the high quality of Kinect2 sensors and the short-range capture, the initial raw reconstruction presents some geometric artifacts, whereas the proposed watertight reconstruction presents a smooth geometry with much fewer artifacts.

Fig. 6(b) shows the reconstruction results with color information. From left to right, the initial raw data are compared with the Poisson volumetric reconstruction method [16] (resolution $2^r \times 2^r \times 2^r$, $r = 7$) and the employed volumetric reconstruction, all with color-per-vertex information without weighted combination of the colors. As can be seen, the employed method presents similar results with the Poisson reconstruction method [16], although it is much faster, as described later in this section. At the right of Fig. 6(b), the final rendered reconstruction, which uses weighted texture blending, is given. The color artifacts are much fewer, the colors are smoothly blended, and the texture is sharper. We highlight that in Fig. 6(b) and all subsequent figures, the light-gray regions (e.g., at the hairs of Argyris) correspond to untextured regions, since some reconstructed vertices are not visible to any camera.

In Fig. 7, the proposed volumetric reconstruction is qualitatively compared with a Truncated Signed Distance Function (TSDF)-based reconstruction [12], [21], at the same volume resolution ($2^r \times 2^{r+1} \times 2^r$, $r = 7$). It is evident that the proposed FT-based method can efficiently handle the depth-measurement noise compared with TSDF that additionally does not produce watertight reconstruction.

*b) Giorgos sequence:* In contrast to the previous example, in this sequence, one can observe color-mismatch problems between the cameras due to change in the lighting conditions. Comparing Kinect1 with the Kinect2 RGB camera, the problem is less frequently observed. However, an example is presented in Fig. 8 to showcase the performance of the employed color-correction method (Section III-D), as well as the importance of the proposed weighted texture blending. As shown in the middle of Fig. 8, the situation improves after color-correction application, whereas the artifacts completely fade out with the weighting of the textures (right).

*c) Stavroula sequence:* A color-correction example with Kinect1 data is provided in Fig. 9. Stavroula was captured with five Kinects1 at distances approx. 2.5 m. The improvement after the application of color correction is visible. One can notice in Fig. 9 (left) the noisy nature of the input Kinect1 data due to multi-Kinect interference.

*2) Medium-Area Configuration (Traditional Sport Skills):* The reconstruction of traditional sport performances is considered in the current subsection. The capturing setup consists of four Kinect2 sensors, placed on a circle of radius close to 4 m. The athletes perform fast sport skills within a large area, sometimes at the distance limits of Kinect2.

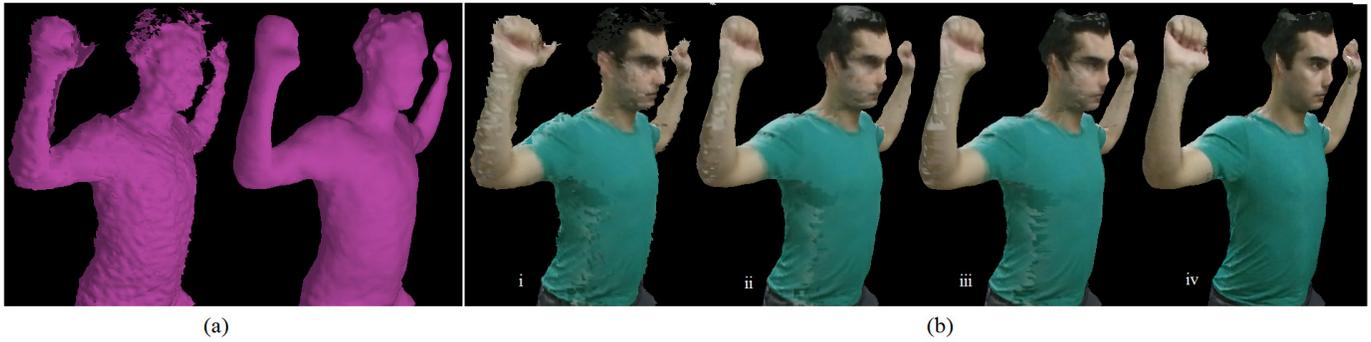

Fig. 6. Argyris sequence. (a) Initial Kinect data, i.e., four separate meshes [13] versus proposed watertight reconstructed geometry. (b) Results with color. i) Four separate meshes. ii) Poisson reconstruction [16] (resolution $2^r \times 2^r \times 2^r$). iii) Proposed watertight volumetric reconstruction (resolution $2^r \times 2^{r+1} \times 2^r$), all with color-per-vertex information without weighted combination of the colors, i.e., equal weights are used. iv) Using weighted blending of the RGB textures, based on the proposed weights. The texture is sharper and the colors are smoothly blended.

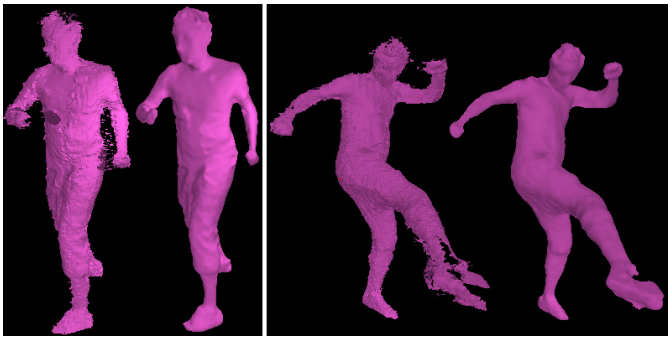

Fig. 7. Argyris sequence. For each pair, the proposed watertight reconstruction (right) is compared with TSDF-based reconstruction (left), at the same volume resolution.

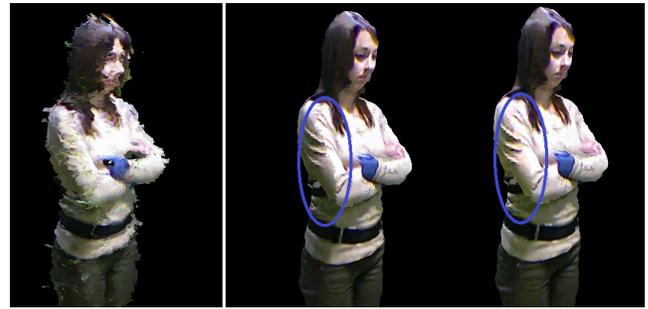

Fig. 9. Stavroula sequence (Kinect1 data). Left: raw reconstruction (five separate meshes). Right: effect of color correction: without and with color correction. In both cases, weighted texture blending was applied.

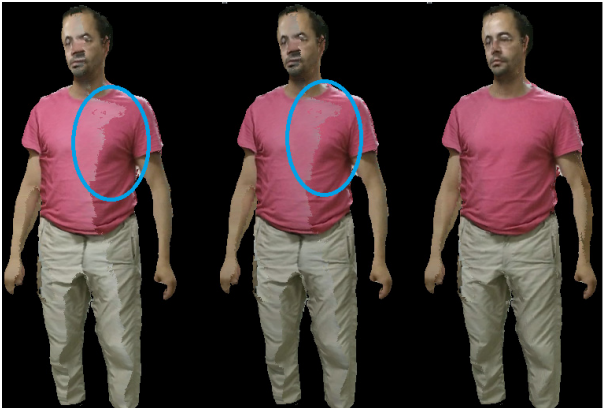

Fig. 8. Giorgos sequence. The effect of color-correction and weighted texture blending. From left to right: initial, after color correction, and after weighted blending.

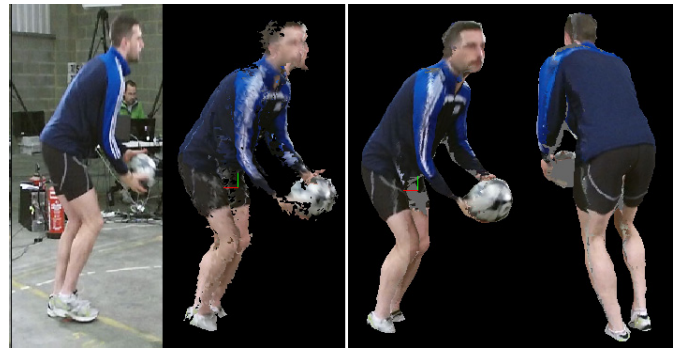

Fig. 10. Gaelic football punt kick. From left to right: original RGB view, raw reconstruction (four separate meshes), and proposed reconstruction from two viewpoints.

*a) Gaelic football punt kick:* Fig. 10 shows an example 3D reconstruction of an athlete during the execution of a Gaelic football skill. Due to the lack of perfect synchronization and the relatively fast motion (notice that the motion blur is visible even in the original view), the captured data are not perfectly aligned. However, the method reconstructs a good-shaped model, whereas the texture weighting method reduces the artifacts significantly.

Fig. 11 shows the positive effect of the smoothing and confidence-based weighting in (1), especially at the separate meshes' boundaries, where noisy input point positions and normals may introduce artifacts.

*b) Gaelic football overhead catch:* The reconstructed 3D geometry of the proposed method is compared with the originally captured data (four separate meshes) in Fig. 12 (left). The volume resolution here is $2^r \times 2^{r+1} \times 2^r$ with $r = 6$. As can be seen, due to the low resolution of the voxel grid, some details, e.g., the hands, are lost. Additional reasons are: 1) the low density of the input captured 3D points (the athlete is far from the cameras) and 2) the non-perfect synchronization, which causes data to be not perfectly aligned and opposite surfaces to cancel out each other.

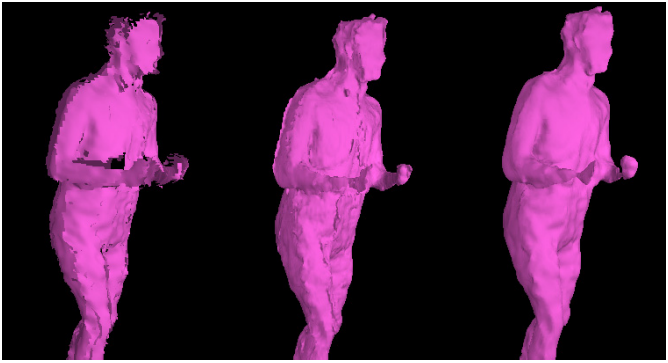

Fig. 11. Gaelic football. Left: raw reconstruction. Middle: reconstruction without the smoothing and confidence-based weighting in (1). Right: reconstruction with the smoothing and confidence-based weighting in (1).

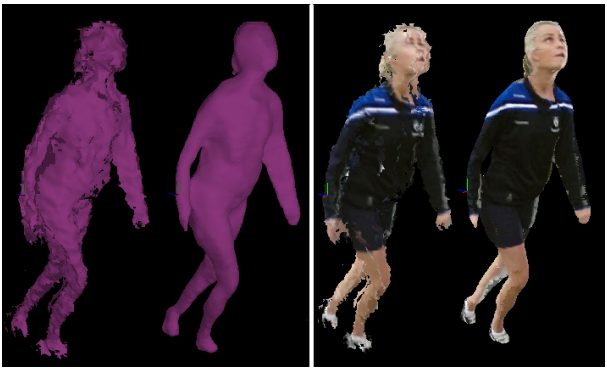

Fig. 12. Gaelic football overhead catch. Left: raw reconstruction versus proposed reconstruction (geometry only). Right: raw reconstruction versus proposed reconstruction, with weighted UV-texture mapping.

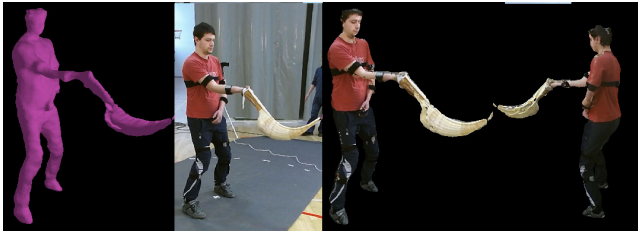

Fig. 13. Jai Alai backhand shot. From left to right: reconstructed geometry, original view, and textured mesh, rendered from two viewpoints.

In Fig. 12 (right), the corresponding final UV-textured model is shown versus the originally captured data.

*c) Traditional Basque sports:* In Fig. 13, an athlete is reconstructed performing a traditional Basque sport skill. Despite the large capture distance and the relatively fast motion, the 3D reconstruction method captures acceptably well the shape and appearance of the athlete.

An additional example is given in Fig. 14 with a female athlete in a fast skill. In this case, thin structures, such as the arms, are not well reconstructed due to very fast motion under nonperfect synchronization conditions. This example reveals the limitation of the capturing system in very fast movements, as any multicamera system without external hardware-based triggering synchronization. This limitation dictates directions for our future work, as will be discussed in Section VI.

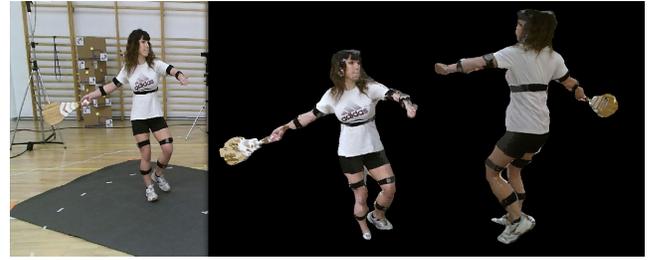

Fig. 14. Pala straight-arm side shot. Original view versus rendered reconstruction from two viewpoints. Due to the fast motion under nonperfect synchronization conditions, thin structures like the arms are not well reconstructed.

TABLE I
COMPARATIVE RESULTS: AVERAGE TIME (ms) OF THE VOLUMETRIC METHODS FOR THE ARGYRIS SEQUENCE

| Vol. Recon. Method | Reconstruction resolution | | |
|---|---|---|---|
| | $r=5$ | $r=6$ | $r=7$ |
| Vol. FT-based proposed (simple) | 10 msec | 22 msec | 102 msec |
| Vol. FT-based proposed | 17 msec | 27 msec | 163 msec |
| Vol. Poisson [16] | 385 msec | 1061 msec | 4602 msec |
| Vol. TSDF-based [12], [21] | 10 msec | 21 msec | 89 msec |

*3) Reconstruction Time/Rate:* The proposed GPU volumetric reconstruction was applied for voxel-grid resolutions $2^r \times 2^{r+1} \times 2^r$, with $r = 5, 6$, and 7. Similarly, a TSDF-based reconstruction was employed using the optimized GPU implementation of the Point-Cloud library (ver.1.8.0, http://pointclouds.org/). Finally, the Poisson reconstruction method [16] was applied with a tree-depth equal to $r + 1$, which corresponds to the same voxel-grid resolution, halved along $Y$. The average number of vertices produced by the proposed method at $r = 7$ is 90k vertices, whereas the corresponding number for Poisson reconstruction is approximately half.

Table I provides the mean execution time results for the proposed volumetric FT-based reconstruction method versus the Poisson method and the TSDF method, considering the Argyris sequence. The experiments ran on a PC with an i7 processor (3.2 GHz), 8-GB RAM, and a CUDA-enabled NVidia GTX 560. As shown in the third row of Table I, the mean reconstruction time for the CPU Poisson method is above 4 s at $r = 7$, whereas the GPU implemented (weighted) FT-based method requires 163 ms, as given in the third row of Table I. The corresponding number for the simple version of the method [without the weighting scheme in (1)] is 102 ms. Therefore, for TI applications in the real-time mode, the simple reconstruction version is used, to increase the reconstruction rate. Compared with the optimized GPU TSDF reconstruction, the proposed method can run at similar time, while producing superior results, as shown in Fig. 7.

Considering all the steps of the reconstruction framework, given in Table II, the total reconstruction time is 167 ms at $r = 7$, which results into near real-time frame rates. The corresponding number for $r = 6$ is 64 ms (15.6 frames/s),

TABLE II
AVERAGE RECONSTRUCTION TIME (ms) AND
RATES FOR ARGYRIS SEQUENCE

| Method's step | Reconstruction resolution | | |
|---|---|---|---|
| | $r = 5$ | $r = 6$ | $r = 7$ |
| Raw point-normal reconstruction | 15 msec | | |
| Calcul. of confidence weights | 4 msec | | |
| Vol. FT-based proposed (simple) | 10 msec | 22 msec | 102 msec |
| Other (e.g. texture mapping) | 5 msec | 23 msec | 46 msec |
| Total (msec) | 34 msec | 64 msec | 167 msec |
| Rate (fps) | 29.4 fps | 15.6 fps | 6.0 fps |

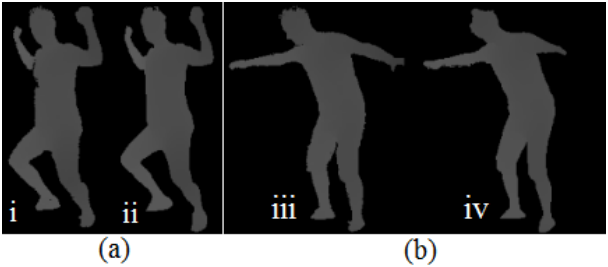

Fig. 15. For each pair (a.i) and (b.iii) originally captured depth, serving as ground truth, and (a.ii) and (b.iv) the reconstructed one.

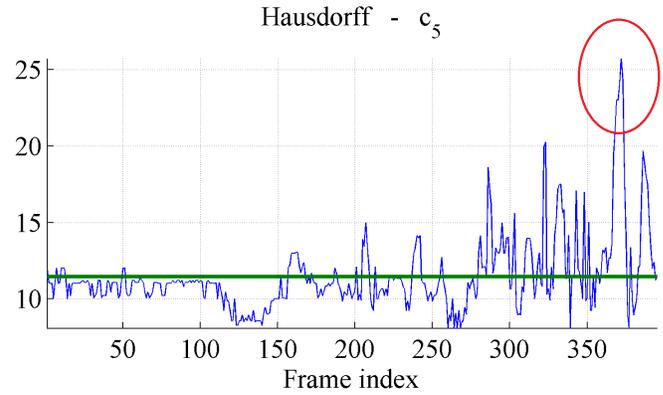

Fig. 16. Argyris sequence: Hausdorff distance $H_5$ ($\downarrow$) along time.

which is quite satisfactory for TI applications with consumer-grade equipment.

### B. 3D Reconstruction—Quantitative Evaluation

The experimental results presented here were extracted from the Argyris sequence. Additional results are given in the Supplementary Material. The objective is threefold: 1) to highlight some practical limitations of the quantitative evaluation methodology of Section III-E; 2) to showcase its validity; and 3) obviously evaluate the employed 3D reconstruction method, presenting also comparative results.

Fig. 15 shows the two examples of reconstructed depth maps versus the corresponding captured maps. As explained in Section III-E, such pairs constitute the input to the proposed evaluation metrics. From Fig. 15(a), one can notice that the reconstructed silhouette (a.ii) is well shaped and smooth, while the corresponding ground-truth (captured) silhouette (a.i) is noisy by nature since it comes from sensor data. This means that the evaluation method is practically limited by the nonperfect ground-truth assumption. More importantly, theoretically the evaluation methodology assumes perfect temporal synchronization and calibration of the sensors. In practice, not perfectly synchronized data from multiple sensors and/or small registration misalignments will lead to worse performance metric values. In other words, the evaluation method addresses the capturing-reconstruction process as a whole. If the capturing process is noisy, the method's capability to differentiate a good reconstruction method from a bad one is reduced. However, as demonstrated in the following, the method presents meaningful results.

In order to assist the reader, the symbols ($\uparrow$) or ($\downarrow$) are used in all subsequent figures to highlight whether a higher or lower metric value, respectively, reflects better performance. The evolution of the Hausdorff distance metric along time, when the sensor $c_5$ is employed for ground truth, is shown in Fig. 16. During the first half of the sequence, the metric remains in low levels, whereas it increases during the second half due to fast motion of the user and loose inter-Kinect synchronization. Some strong peaks, as the highlighted one, correspond to missing-limb cases, as shown in Fig. 15(b). In accordance to Section III-E1, the Hausdorff distance in this case is equal to the length of the missing limb that is approximately 25 pixels.

Fig. 17 shows the mean values (considering all ground-truth views $c_k$) for the metrics that reflect the volume/geometry reconstruction quality. The results are given for voxel-grid resolutions $r = 5$, 6, and 7, considering the employed and the Poisson reconstruction method. All metrics decrease as the reconstruction resolution increases, as expected. In all plots, the employed reconstruction method presents similar or slightly better performance than the Poisson method. This is explained by the doubled resolution along $Y$ for the employed method. Only for low resolution ($r = 5$), and according to VRE and Hausdorff distance, the Poisson method performs better.

The Hausdorff distance for two views ($c_0$ and $c_2$) is given in Fig. 18. The same conclusions can be drawn. An additional conclusion is that the metric values for view $c_2$ are higher, as expected, since sensor $c_2$ does not participate in the reconstruction process.

Finally, the results with respect to the RGB appearance quality are given in Fig. 19, using the structure similarity index (WMS3IM) metric. The color-per-vertex representation approach is compared with the UV-texture-mapping approach, considering three reconstruction resolutions. The results are meaningful, since: 1) WMS3IM improves as the resolution increases; 2) the color-per-vertex representation always performs worse since it produces blurred (lower resolution) rendered views than the originally captured one; and 3) on the other hand, the UV-texture-mapping approach performs well, even at $r = 5$, since it directly maps the high-definition captured RGB textures. Finally, the WMS3IM values in Fig. 19 (left) are higher than those in Fig. 19 (right), since sensor $c_0$ participates in the reconstruction and texture-mapping process.

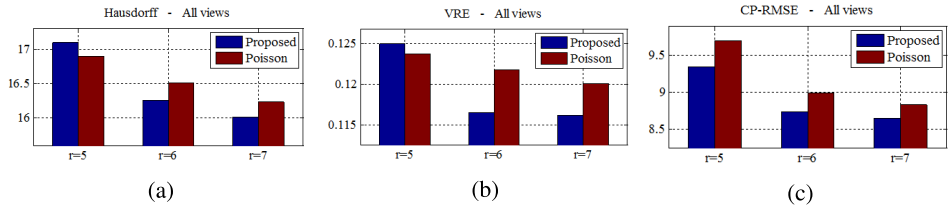

Fig. 17. Argyris sequence. Reconstruction performance, considering the mean for all views. The results for the employed and the Poisson method are given with respect to reconstruction resolution. (a) Hausdorff distance ($\downarrow$). (b) VRE measure ($\downarrow$). (c) CP-RMSE measure ($\downarrow$).

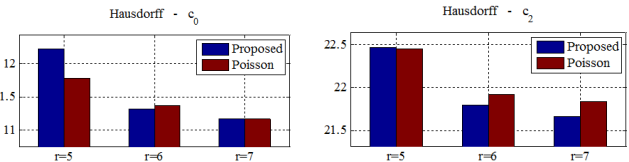

Fig. 18. Argyris sequence. Hausdorff distances $H_k$ ($\downarrow$) considering sensors $c_k, k = \{0, 2\}$.

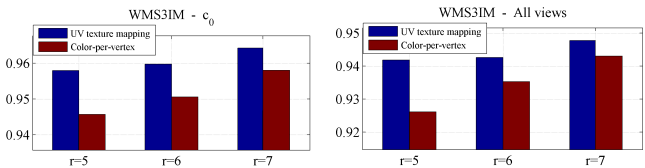

Fig. 19. Argyris sequence. WMS3IM similarity index ($\uparrow$) considering sensor $c_0$ (left) and all sensors (right). The results for the color-per-vertex representation and UV-texture mapping are given.

### C. Volume-Based Motion Tracking

Our motion capturing system is evaluated mainly using a data set of Gaelic and Basque traditional sports provided by the project RePlay. The specific data set was selected for experimentation because, apart from multiple Kinect skeleton data, Vicon marker-based ground truth is available. The 15-joint skeleton structure, extracted by the proposed method, constitutes a subset of the Kinect and Vicon structures, and therefore there exists one-to-one joint correspondences between the three structures. The captured motions are challenging and fast, with severe self-occlusions and simultaneous movements of several body parts. Sequences from different sport skill captures were chosen, characterized by short, quick movements. The data used in the experiments can be found at http://vcl.iti.gr/3dTI/TCSVT/dataset.

An illustrative skeleton-tracking example is given in Fig. 20. As can be seen at the top of Fig. 20, the estimates of the proposed method may be inaccurate at the presence of large reconstructed objects (e.g., the ball) touching the human limbs. This limitation is expected to be overridden by fusing in our method data from an inertia measurement unit. The plot diagram at the bottom of Fig. 20 shows the estimated anthropometric angle (between two bones) along time for the most important limb of this skill. The valley of the curves at the beginning of the sequence corresponds to the flexion of the knee for kicking. Where Kinect2 loses tracking for a few frames after the fast knee flexion (large valley that reaches 0°), the proposed method tracks well the motion.

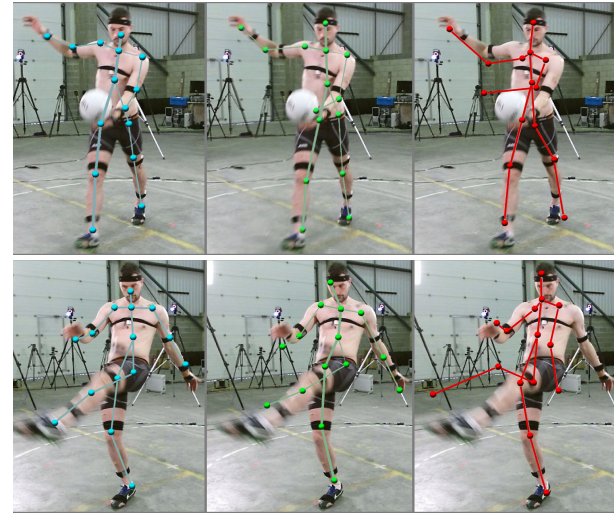

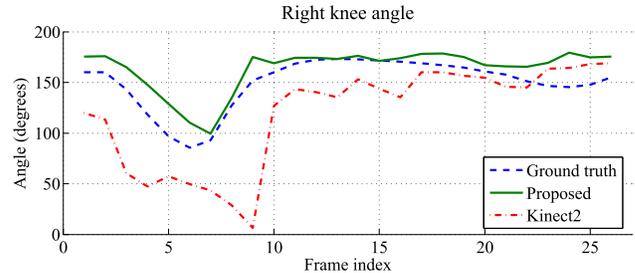

Fig. 20. Gaelic football punt kick. Qualitative and quantitative skeleton-tracking results. Ground truth (Vicon) with cyan, proposed with green, and Kinect2 (using the best skeleton among the 4 Kinect2 sensors) with red.

Table III shows the comparisons of the angle estimates with the ground truth, using the RMSE and the mean absolute error ($\downarrow$), while highlighting with bold the most important limb, as analyzed by biomechanical engineers.

*1) Runtime Evaluation:* The experiments ran on a PC with an Intel Core i7 processor at 3.5-GHz, 16-MB RAM, and the NVidia GTX 680 graphics card. The proposed skeleton-tracking method can achieve frame rates higher than 10 frames/s.

1) The human-body volumetric function is reconstructed on the GPU at a volume resolution $r = 6$ within approximately 20 ms (see also Section V-A).
2) The processing time for volume binarization and skeletonization, running on a CPU thread, is 10 ms.
3) Creating and processing the skeletal graph lasts 30 ms.

ALEXIADIS et al.: INTEGRATED PLATFORM FOR LIVE 3D HUMAN RECONSTRUCTION AND MOTION CAPTURING 811TABLE III
SKELETON-TRACKING RESULTS: MEAN ERRORS BETWEEN
THE ESTIMATED ANGLE AND THE GROUND TRUTH

| Sequences | Left Elbow | | Right Elbow | | Left Knee | | Right Knee | |
|---|---|---|---|---|---|---|---|---|
| **Kinect2** | MAE | RMSE | MAE | RMSE | MAE | RMSE | MAE | RMSE |
| Jai Alai (Sidearm Shot) | 31.19 | 35.69 | **25.96** | **38.35** | 22.58 | 30.68 | 12.99 | 18.22 |
| Pelota (R-H Serve) | 40.35 | 48.04 | **29.30** | **38.92** | 3.43 | 8.34 | 14.88 | 20.04 |
| Handball (R-H Volley) | 38.62 | 45.71 | **34.06** | **46.54** | 13.71 | 26.06 | 7.74 | 18.38 |
| Gaelic Football (R-F Punt Kick) | 5.39 | 10.42 | 25.084 | 25.79 | 18.34 | 20.53 | **21.03** | **53.11** |
| **Proposed** | | | | | | | | |
| Jai Alai (Sidearm Shot) | 25.29 | 29.34 | **14.26** | **21.60** | 9.940 | 13.68 | 12.87 | 18.11 |
| Pelota (R-H Serve) | 16.79 | 25.59 | **16.39** | **25.09** | 9.30 | 19.41 | 9.04 | 16.29 |
| Handball (R-H Volley) | 17.78 | 26.51 | **14.65** | **23.35** | 7.15 | 17.51 | 10.014 | 25.31 |
| Gaelic Football (R-F Punt Kick) | 16.34 | 26.47 | 23.37 | 31.42 | 15.88 | 17.32 | **10.85** | **17.20** |

4) Fitting a skeleton to the graph requires less than 10 ms. Although the implementation of the method after the volume extraction is not optimized, it allows the skeleton estimation at rates higher than 10 frames/s.

## VI. CONCLUSION AND FUTURE WORK

In this paper, the main elements of an integrated system that targets real-time future 3D applications were described, including multi-Kinect2 capturing and fast 3D reconstruction of moving humans, as well skeleton-based motion tracking from multiple depth cameras. Regarding these elements, novel approaches were proposed and/or the adaptation of existing ones were described. Simultaneously, a novel framework for the quantitative evaluation of 3D reconstruction systems has been proposed.

In terms of research and development, some limitations of the ongoing system have also been discussed. Overriding these limitations is the subject of ongoing research. Regarding the nonperfect synchronization issue with consumer-grade RGB-D sensors, which may deteriorate the reconstruction quality in fast motion, we work toward spatio-temporal interpolation via estimation of the separate 3D data misalignment. With respect to the skeleton-tracking method, the limitations regarding topology change (e.g., piece hands together) are expected to be overriden by a skeleton-fitting scheme, where the limbs of a user-specific skeleton model are fitted to the extracted MST. In addition, by splitting the rigid-body part into upper and lower segments and fusing in our method data from two inertial measurement units, we aim at handling the limitations due to the assumption that the trunk joints move rigidly.

To increase realism, with respect to 3D reconstruction of humans, a generic future work direction is the improvement of the visual quality and frame rates by continuously investigating more efficient solutions. For example, in many applications, the reconstruction of user's face is more important than other body parts, and therefore we investigate toward the real-time deformation and fusion of a prescanned user's head model with the captured 3D data. In real-time applications, such as TI, both: 1) realistic replications of the users appearance (heavy data) and 2) natural interaction among geographically remote user (real-time exchange of the 3D reconstructions among remote locations) are required. The above contradiction also highlights the need to research both in the compression of the 3D replicants and in the network layer to offer novel TI architectures, allowing to scale up the interaction among a large number of users capable of supporting such exciting applications.

## REFERENCES

[1] R. Vasudevan et al., "High-quality visualization for geographically distributed 3-D teleimmersive applications," *IEEE Trans. Multimedia*, vol. 13, no. 3, pp. 573–584, Jun. 2011.

[2] A. Doumanoglou, D. S. Alexiadis, D. Zarpalas, and P. Daras, "Toward real-time and efficient compression of human time-varying meshes," *IEEE Trans. Circuits Syst. Video Technol.*, vol. 24, no. 12, pp. 2099–2116, Dec. 2014.

[3] S. Crowle, A. Doumanoglou, B. Poussard, M. Boniface, D. Zarpalas, and P. Daras, "Dynamic adaptive mesh streaming for real-time 3D teleimmersion," in *Proc. 20th Int. Conf. Web 3D Technol.*, Crete, Greece, 2015, pp. 269–277.

[4] P. Fechteler et al., "A framework for realistic 3D tele-immersion," in *Proc. 6th Int. Conf. Comput. Vis./Comput. Graph. Collaboration Techn. Appl. (MIRAGE)*, Berlin, Germany, 2013, Art. no. 12.

[5] J.-S. Franco and E. Boyer, "Efficient polyhedral modeling from silhouettes," *IEEE Trans. Pattern Anal Mach. Intell.*, vol. 31, no. 3, pp. 414–427, Mar. 2009.

[6] G. Vogiatzis, C. Hernández, P. H. S. Torr, and R. Cipolla, "Multiview stereo via volumetric graph-cuts and occlusion robust photo-consistency," *IEEE Trans. Pattern Anal. Mach. Intell.*, vol. 29, no. 12, pp. 2241–2246, Dec. 2007.

[7] Y. Furukawa and J. Ponce, "Carved visual hulls for image-based modeling," *Int. J. Comput. Vis.*, vol. 81, no. 1, pp. 53–67, 2009.

[8] J. Carranza, C. Theobalt, M. A. Magnor, and H.-P. Seidel, "Free-viewpoint video of human actors," *ACM Trans. Graph.*, vol. 22, no. 3, pp. 569–577, 2003.

[9] E. de Aguiar, C. Stoll, C. Theobalt, N. Ahmed, H.-P. Seidel, and S. Thrun, "Performance capture from sparse multi-view video," *ACM Trans. Graph.*, vol. 27, no. 3, 2008, Art. no. 98.

[10] J. Gall, B. Rosenhahn, T. Brox, and H.-P. Seidel, "Optimization and filtering for human motion capture: A multi-layer framework," *Int. J. Comput. Vis.*, vol. 87, no. 1, pp. 75–92, 2010.

[11] G. Turk and M. Levoy, "Zippered polygon meshes from range images," in *Proc. SIGGRAPH*, 1994, pp. 311–318.

[12] B. Curless and M. Levoy, "A volumetric method for building complex models from range images," in *Proc. SIGGRAPH*, 1996, pp. 303–312.

[13] D. S. Alexiadis, D. Zarpalas, and P. Daras, "Real-time, full 3-D reconstruction of moving foreground objects from multiple consumer depth cameras," *IEEE Trans. Multimedia*, vol. 15, no. 2, pp. 339–358, Feb. 2013.

[14] D. S. Alexiadis, D. Zarpalas, and P. Daras, "Real-time, realistic full-body 3D reconstruction and texture mapping from multiple kinects," in *Proc. IEEE IVMSP*, Jun. 2013, pp. 1–4.

[15] A. Maimone and H. Fuchs, "Real-time volumetric 3D capture of room-sized scenes for telepresence," in *Proc. IEEE 3DTV-Conf., True Vis.-Capture, Transmiss. Display 3D Video*, Oct. 2012, pp. 1–4.

[16] M. Kazhdan, M. Bolitho, and H. Hoppe, "Poisson surface reconstruction," in *Proc. 4th Eurograph. Symp. Geometry Process.*, 2006, pp. 61–70.

[17] M. Kazhdan, "Reconstruction of solid models from oriented point sets," in *Proc. 3rd Eurograph. Symp. Geometry Process.*, 2005, Art. no. 73.

[18] A. Maimone and H. Fuchs, "Encumbrance-free telepresence system with real-time 3D capture and display using commodity depth cameras," in *Proc. IEEE ISMAR*, Oct. 2011, pp. 137–146.

[19] A. Maimone, J. Bidwell, K. Peng, and H. Fuchs, "Enhanced personal autostereoscopic telepresence system using commodity depth cameras," *Comput. Graph.*, vol. 36, no. 7, pp. 791–807, 2012.

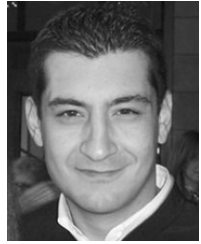

**Dimitrios S. Alexiadis** received the Diploma and Ph.D. degrees in electrical and computer engineering from Aristotle University of Thessaloniki, Thessaloniki, Greece, in 2002 and 2009, respectively.

Since 2011, he has been with the Information Technologies Institute, Centre for Research and Technology–Hellas, Thessaloniki, as a Post-Doctoral Research Associate. From 2006 to 2015, he co-authored 17 high-impact journal papers and more than 25 international conference papers. His current research interests include multimedia processing, stereo/multiview processing, 3D reconstruction and compression, 3D flow estimation, and scene analysis.

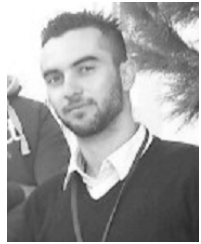

**Anargyros Chatzitofis** received the Diploma degree from the Electrical and Computer Engineering School, National Technical University of Athens, Athens, Greece, where he is currently pursuing the Ph.D. degree in computer science.

Since 2012, he has been a Research Assistant with the Information Technologies Institute, Centre for Research and Technology–Hellas, Thessaloniki, Greece. From 2013 to 2015, he co-authored ten international conference papers. His research interests include data acquisition and processing from several sensors, multimedia, and stereo/multiview processing, focusing on human motion capturing and analysis, by using depth sensors, wireless inertial measurement units, and RGB-cameras.

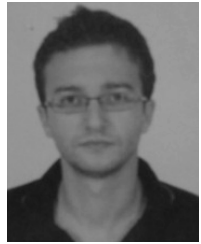

**Nikolaos Zioulis** received the Diploma degree in electrical and computer engineering from Aristotle University of Thessaloniki, Thessaloniki, Greece, in 2012.

He has been a Research Assistant with the Information Technologies Institute, Centre for Research and Technology–Hellas, Thessaloniki, since 2013. His research interests include software design and development, computer vision, motion analysis, 3D graphics, and GPU processing.

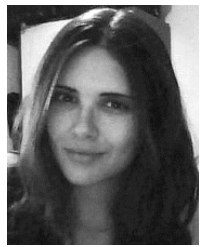

**Olga Zoidi** received the B.Sc. degree in mathematics, the Diploma degree in electrical and computer engineering, and the Ph.D. degree in informatics from Aristotle University of Thessaloniki, Thessaloniki, Greece, in 2004, 2009, and 2014, respectively.

Since 2015, she has been with the Information Technologies Institute, Centre for Research and Technology–Hellas, Thessaloniki, as a Post-Doctoral Research Fellow. Since 2009, she has co-authored one book chapter, six papers in high-impact scientific journals, and 16 papers in international conferences. Her research interests include image and video processing, 3D reconstruction, computer vision, and pattern recognition.

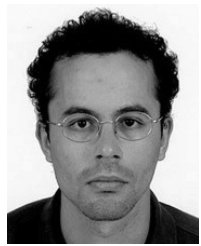

**Georgios Louizis** received the B.Sc. degree from the Department of Informatics, Aristotle University of Thessaloniki (AUTh), Thessaloniki, Greece, in 2001.

He has been a Research Assistant with the Artificial Intelligence and Information Analysis Laboratory, AUTh, the Signal Image and Multimedia Processing Laboratory, University of British Columbia, Vancouver, BC, Canada, and AquaMed ME, Doha, Qatar. Since 2013, he has also been a Research Assistant with the Information Technologies Institute, Centre for Research and Technology–Hellas, Thessaloniki.

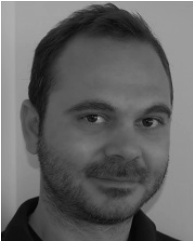

**Dimitrios Zarpalas** received the Diploma degree in electrical and computer engineering from Aristotle University of Thessaloniki (AUTh), Thessaloniki, Greece, the M.Sc. degree in computer vision from The Pennsylvania State University, State College, PA, USA, and the Ph.D. degree in medical informatics from the School of Medicine, AUTh.

Since 2008, he has been an Associate Researcher with the Information Technologies Institute, Centre for Research and Technology–Hellas, Thessaloniki, where he has been involved in research on real-time tele-immersion through 3D reconstruction from multiple sensors, 3D shape and motion analysis, and 3D shape descriptor extraction. He has co-authored 46 papers in peer-reviewed international journals, conference proceedings, and books.

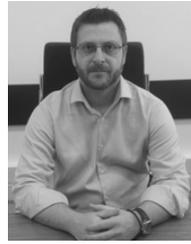

**Petros Daras** (SM'13) received the Diploma degree in electrical and computer engineering, the M.Sc. degree in medical informatics, and the Ph.D. degree in electrical and computer engineering from Aristotle University of Thessaloniki, Thessaloniki, Greece, in 1999, 2002, and 2005, respectively. He is currently a Principal Researcher with the Information Technologies Institute, Centre for Research and Technology–Hellas, Thessaloniki, Greece. He has co-authored more than 40 papers in refereed journals, 29 book chapters, and more than 100 papers in international conferences. His research interests include multimedia processing, multimedia and multimodal search engines, 3D reconstruction from multiple sensors, dynamic mesh coding, medical image processing, and bioinformatics.